\pdfoutput=1

\documentclass[11pt]{article}

\usepackage{acl}

\usepackage{times}
\usepackage{latexsym}

\usepackage[T1]{fontenc}

\usepackage[utf8]{inputenc}

\usepackage{microtype}

\usepackage{inconsolata}

\usepackage{graphicx}
\usepackage{multirow}
\usepackage{csquotes}
\usepackage{caption}
\usepackage{fancyvrb}
\usepackage{tcolorbox}
\usepackage{float}

\usepackage{tikz}
\usepackage{circledsteps}

\newcommand{\whatif}{{\large \textsc{what-if}}}
\newcommand*\circled[1]{\tikz[baseline=(char.base)]{
            \node[shape=circle,draw,inner sep=1pt] (char) {#1};}}

\title{{\textsc{\LARGE what-if}}: Exploring Branching Narratives by Meta-Prompting Large Language Models}
 
\author{Runsheng "Anson" Huang \\
  University of Pennsylvania \\
  \texttt{rhuang99@seas.upenn.edu} \\\And
   Lara J. Martin\\
  University of Maryland,  \\
  Baltimore County\\
  \texttt{laramar@umbc.edu} \\\And
  Chris Callison-Burch \\
  University of Pennsylvania \\
  \texttt{ccb@seas.upenn.edu } \\}

\begin{document}
\maketitle
\begin{abstract}
\textbf{WHAT-IF}---\textbf{W}riting a \textbf{H}ero's \textbf{A}lternate \textbf{T}imeline through \textbf{I}nteractive \textbf{F}iction---is a system that uses zero-shot meta-prompting to create branching narratives from a prewritten story. Played as an interactive fiction (IF) game, \whatif~lets the player choose between decisions that the large language model (LLM) GPT-4 generates as possible branches in the story. Starting with an existing linear plot as input, a branch is created at each key decision taken by the main character. By meta-prompting the LLM to consider the major plot points from the story, the system produces coherent and well-structured alternate storylines. \whatif~stores the branching plot tree in a graph which helps it to both keep track of the story for prompting and maintain the structure for the final IF system. 
A demo of \whatif~can be found at \url{https://what-if-game.github.io/}.
\end{abstract}

\section{Introduction}
Interactive fiction (IF) is a form of text-based story  where a user performs some action that moves the plot along. 
From Choose Your Own Adventure\texttrademark~books to open-world video games, branching narratives are an interesting element of interactive fiction that enables the reader to change the outcome of the narrative.

\begin{figure}[t!]
\includegraphics[width=7.6cm]{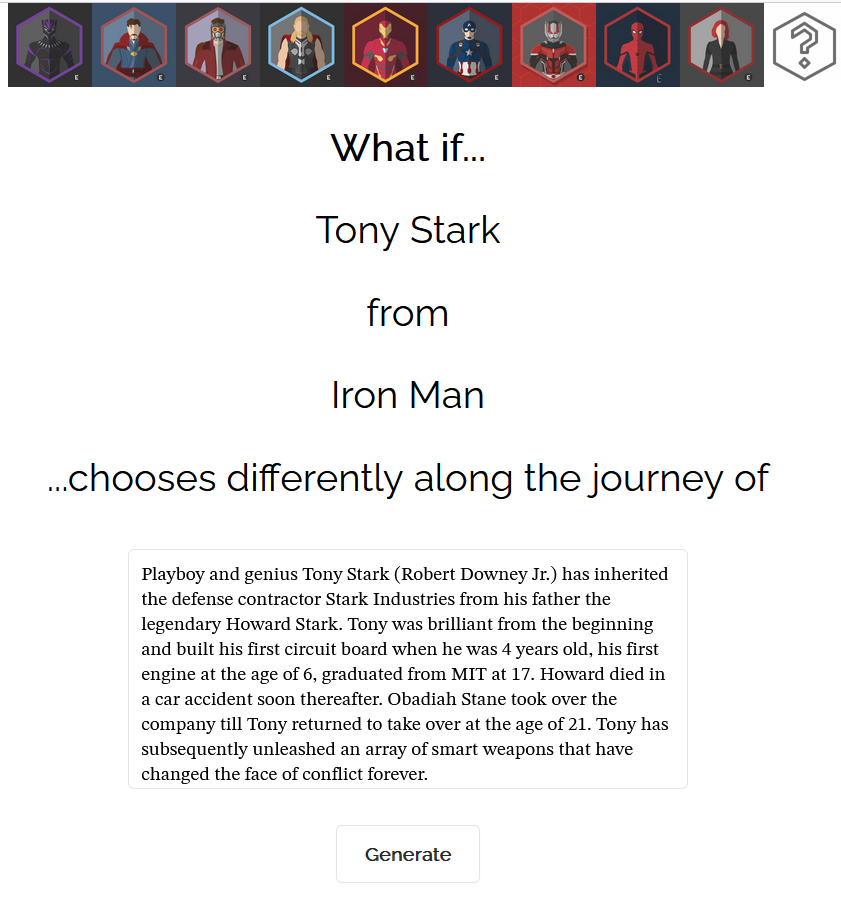}
\caption{The \whatif~user interface, filled with the main character, title, and the plot of Marvel's  \textit{Iron Man} taken from \href{https://www.imdb.com/title/tt0371746/plotsummary/}{IMDb}. The main character, title, and plot summary can all be edited by the user.}
\label{fig:UI}
\end{figure}

In this paper, we present \textbf{WHAT-IF} (\textbf{W}riting a \textbf{H}ero's \textbf{A}lternate \textbf{T}imeline through \textbf{I}nteractive \textbf{F}iction), a system for controllable, branching narrative generation in English. 
Starting with a human-written plot, \whatif~produces distinct changes in the plot by considering alternate actions the main character could have taken. We create these choice points by having a large language model (LLM) generate a new sub-goal for the character, which significantly changes subsequent plot points in the story.
We use this LLM-driven branching narrative to create an interactive fiction system (Fig. \ref{fig:UI}) allowing the user to choose which branches to follow in the story tree. 

In contrast with previous LLM-driven interactive fiction systems like AI Dungeon \cite{AIDungeon} which rely entirely on the user to move the story along,
\whatif~generates more constrained stories. The user has a degree of control, but the stories generated by the LLM maintain a high degree of thematic consistency with the original story, even with many possible endings.

\begin{figure}[h]
\includegraphics[width=7.6cm]{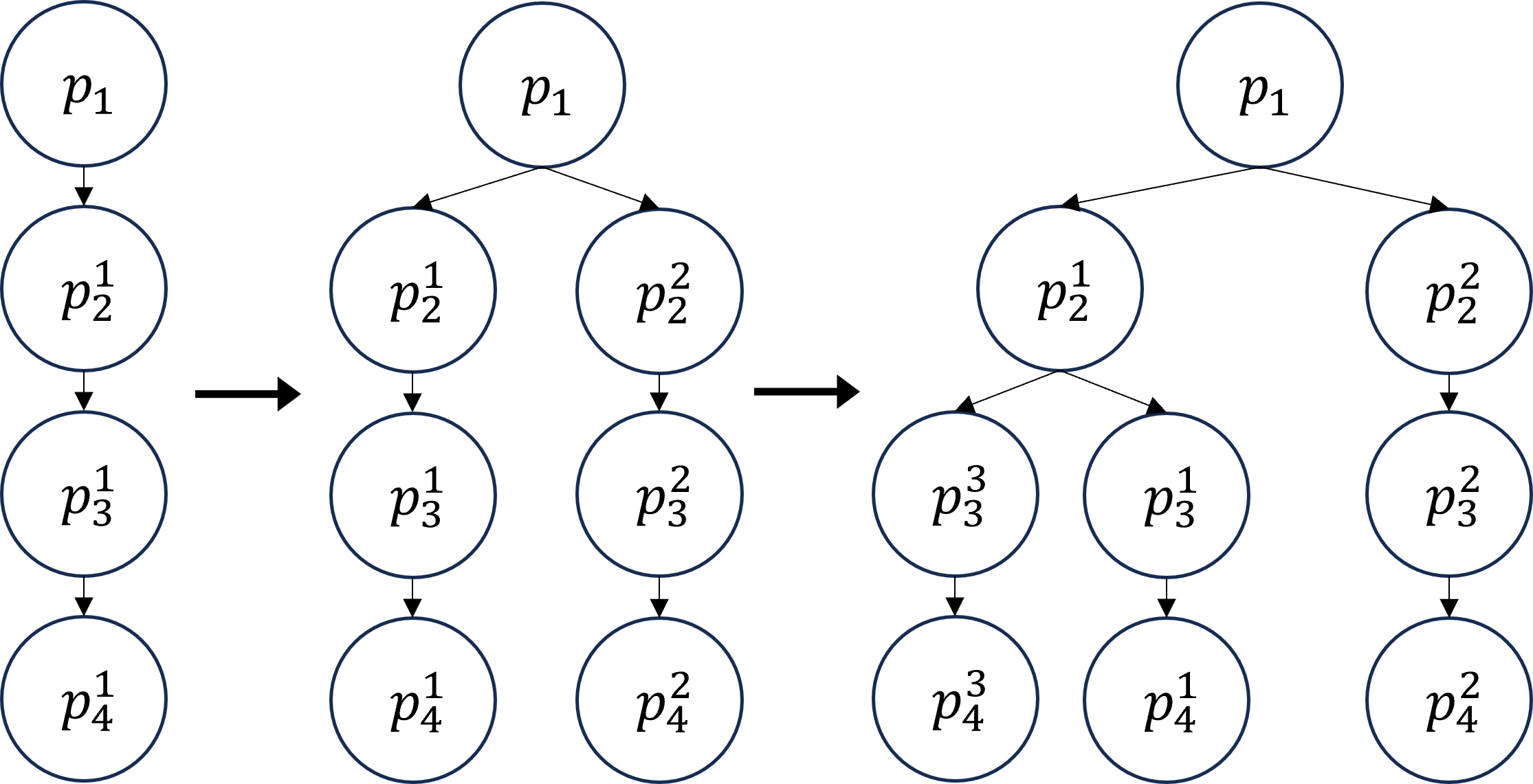}
\caption{An illustration of the branching plot. The human-written plot $p^1$ is extracted and turned into the initial tree. Then \whatif~uses GPT-4 to recursively add alternate plots until a full binary tree is created.}
\label{fig:branch}
\end{figure}

Structure can help LLMs better understand stories \cite{Dong_etal_2023}.
\whatif~breaks down the narrative into events so that the structure of the plot can be tracked throughout the generation. 
A branching plot structure, such as that shown in Figure \ref{fig:branch}, is generated before the user plays the game. Each branch of the story will lead to an ending.

As each story requires asking different questions to generate hypothetical alternate plots, we also employ meta-prompting. That is, the system generates story-specific prompts to be fed into another instance of the LLM to answer.

Although we present the system as an interactive fiction game, we believe that \whatif~has potential applications for helping authors explore alternative paths in storylines they are writing. For example, LLMs have been used as writing aids \cite{Swanson_etal_2021,Yuan_etal_2022_Wordcraft,CALYPSO,Gero_Long_Chilton_2023}, and have even been shown to help people author IFs \cite{Park_Shin_Kim_Bae_2023}.

In the rest of this paper, we will discuss related work about using LLMs for interactive fiction and story generation, describe the architecture of \whatif, and then give example narratives that were generated. The code for generating branching narratives with \whatif~can be found at \url{https://github.com/what-if-game/what-if-game.github.io}.

\section{Related Work}
\subsection{Storytelling with LLMs}
Story generation has a long history in classical planning \cite{Meehan_1977,Lebowitz_1984,Young_etal_2013}, where the resulting story comes from the plan extracted from a graph of plot points.
These systems enable long-term coherent story plots to form, but have no capabilities for text generation.

On the other hand, LLMs yield flexible, grammatical natural language generation.
While early LLMs had mixed results with story generation \cite{See_etal_2019}, more modern LLMs are telling better stories and are able to generate short stories at a human level \cite{xie-etal-2023-next,Wang_2025}.

There has been plenty of work on controlled story generation with LLMs \cite{Lal_Chambers_Mooney_Balasubramanian_2021,razumovskaia-etal-2024-little,Yang_etal_2022,Qin_Zhao_2022,Zhu_etal_2023_End,Kumaran_2024,Wang_2025,Xia_2025,Zimmerman_2025}. 
Most of this work consists of higher-level ``planning'' with LLMs, where an outline or set of keywords is expanded into a longer-form story.
There have also been recent works using retrieval-augmented LLMs for story generation \cite{CALYPSO,wen-etal-2023-grove}.
However, people have only just begun to explore mixing more traditional AI planning techniques with LLMs, namely by generating components that would be fed into a planner or using pre-conditions and effects in a similar manner to a planner \cite{Martin_2021,Simon_Muise_2022,Kelly_etal_2023,Ye_etal_2023,Farrell_2024}.

LLMs have also been used for story-generation--related tasks, including generating story worlds \cite{Johnson-Bey_etal_2023}, doing story sifting \cite{Méndez_Gervás_2023}---that is, creating a story from a collection of facts---and generating stories via simulation of characters \cite{Yu_2025}.

\subsection{Interactive Narrative Generation}

There have been \textit{interactive} narrative systems using planning methods as well \cite{Riedl_Bulitko_2013,Sanghrajka_2019,Ware_Siler_2021}.
So far, neural IF work has included tasks such as generating interactive fiction worlds \cite{Ammanabrolu_etal_2020_bringing,Zhou_2025}, scenes \cite{Kumaran_etal_2023}, branching quests \cite{DeLima_Feijo_Furtado_2021}, or choices for the user \cite{Harmon_Rutman_2023}. We take inspiration from \citet{DeLima_Feijo_Furtado_2021} and adapt their quest tree planning methods to generating a plot tree with an LLM.

There has been a collection of work looking at interactive fiction through roleplaying as well, using recurrent language models \cite{Louis_Sutton_2018}, reinforcement learning \cite{Martin_Sood_Riedl_2018,Ammanabrolu_etal_2021}, and LLMs \cite{ccb2022dungeons,Ashby_Webb_Knapp_Searle_Fulda_2023,Wang_etal_2023,Shao_Li_Dai_Qiu_2023,Yan_Li_Zhang_Wang_Yang_Yan_2023}.

\begin{figure}[ht]
\includegraphics[width=7.6cm]{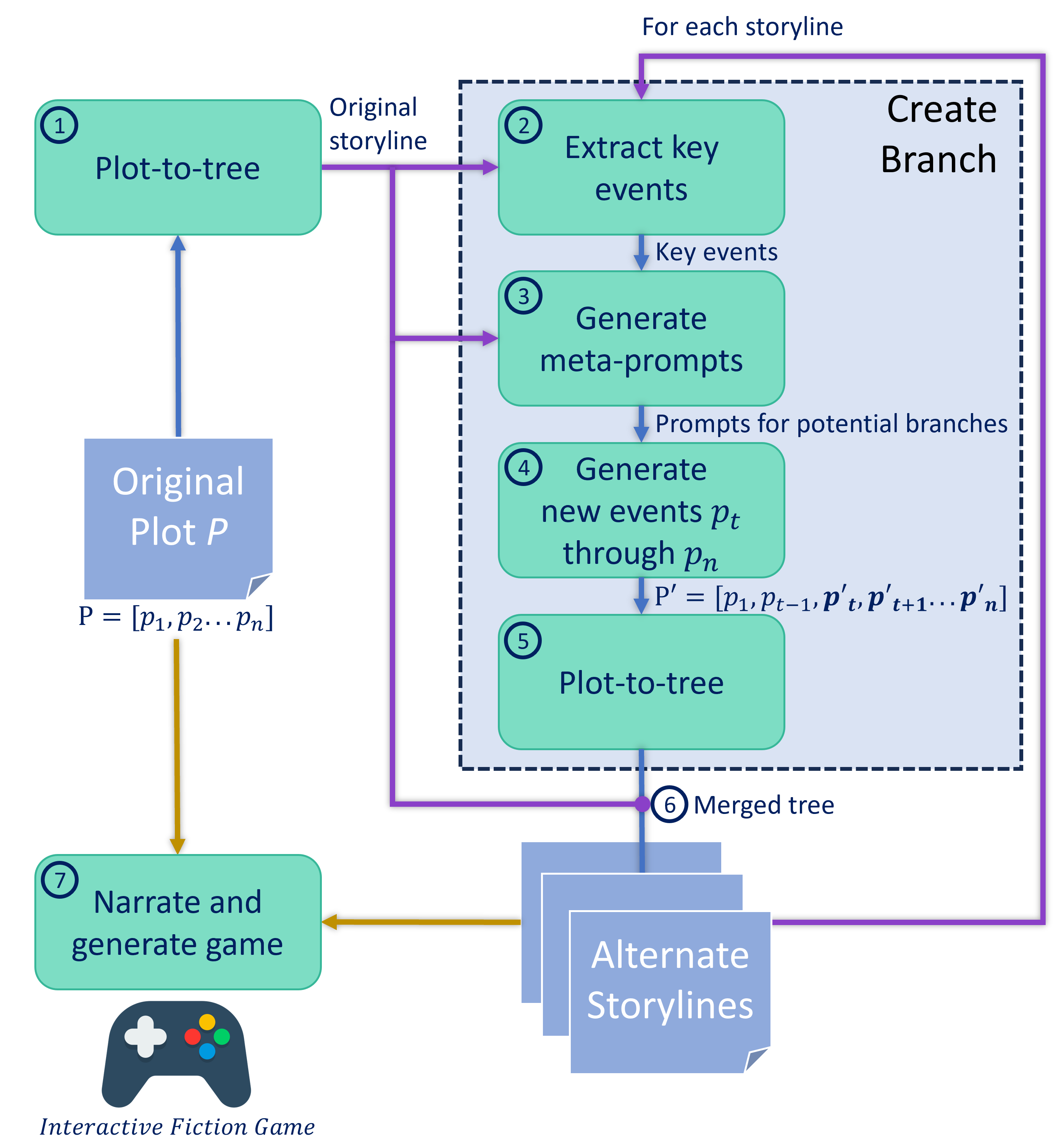}
\caption{\whatif~code structure.  \textcircled{1} The original plot initializes the branching plot tree, which is fed to the branch creation to \textcircled{2} extract key events and \textcircled{3} generate meta-prompts. \textcircled{4} The prompts are then used to generate the new events that branch from the current time point of the story. \textcircled{5} The new events are turned into their own plot tree and \textcircled{6} are merged with the existing tree as a new branch. \textcircled{7} The final tree is converted into alternate storylines which are used to narrate the game.}
\label{fig:diagram}
\end{figure}

There are several systems that are of note as comparisons to our system \whatif.
\citet{Patel_etal_2024}'s SWAG story generation system uses an LLM to both generate actions (story events) and then select the best action using Direct Preference Optimization (DPO) \cite{DPO}. Although we are not selecting the action---the user is, we are relying on the LLM to generate story branches. However, instead of finetuning on pre-written stories, \whatif~uses zero-shot prompting to generate new events.

Similarly, like \citet{Harmon_Rutman_2023}'s works, we use LLMs to generate branching narratives by exploring character choices. However, instead of using story-specific questions as part of the prompt, \whatif~uses meta-prompts so that the model can ask itself story-specific questions to aid in generation.

\section{System Architecture}
\whatif~(Fig. \ref{fig:diagram}) consists of five major phases: initializing the branching plot tree (\S\ref{sec:init}), extracting key events (\S\ref{sec:key_event}), generating meta-prompts (\S\ref{sec:prompts}), branching \& merging the tree (\S\ref{sec:branching}), and narrating the game (\S\ref{sec:narrating}). These components
ensure the system chooses logical branching points, maintains a coherent and consistent story structure in alternate storylines, and creates an immersive playing experience.

For each component of the pipeline, we use GPT-4, which was the state-of-the-art at the time of creating \whatif. Each instance of the model was used out-of-the-box (not finetuned) to test how well it could be pushed to generated well-formed alternative storylines for stories and characters it has most likely memorized. This is also the reasoning behind choosing well-known stories.

\subsection{Initializing the Branching Plot Tree}
\label{sec:init}
Most interactive fiction games tell the story of the protagonist (the player) embarking on a journey full of challenges, making key decisions, and reaching their goals along the way. The major challenge for human interactive fiction writers when designing branching narratives is to propose meaningful key decisions that offer players agency in shaping the story. Inspired by \citet{DeLima_Feijo_Furtado_2021}'s previous work of adopting a quest tree in story planning and branching, we create a \textbf{branching plot tree} structure to capture key decisions in a given story plot as branching points. Here, we define a \textbf{key decision} as the decision a main character takes in response to a given state (obstacles, dangers, etc.) which influences how the rest of the story unfolds.

\begin{figure}[h]
\includegraphics[width=7.6cm]{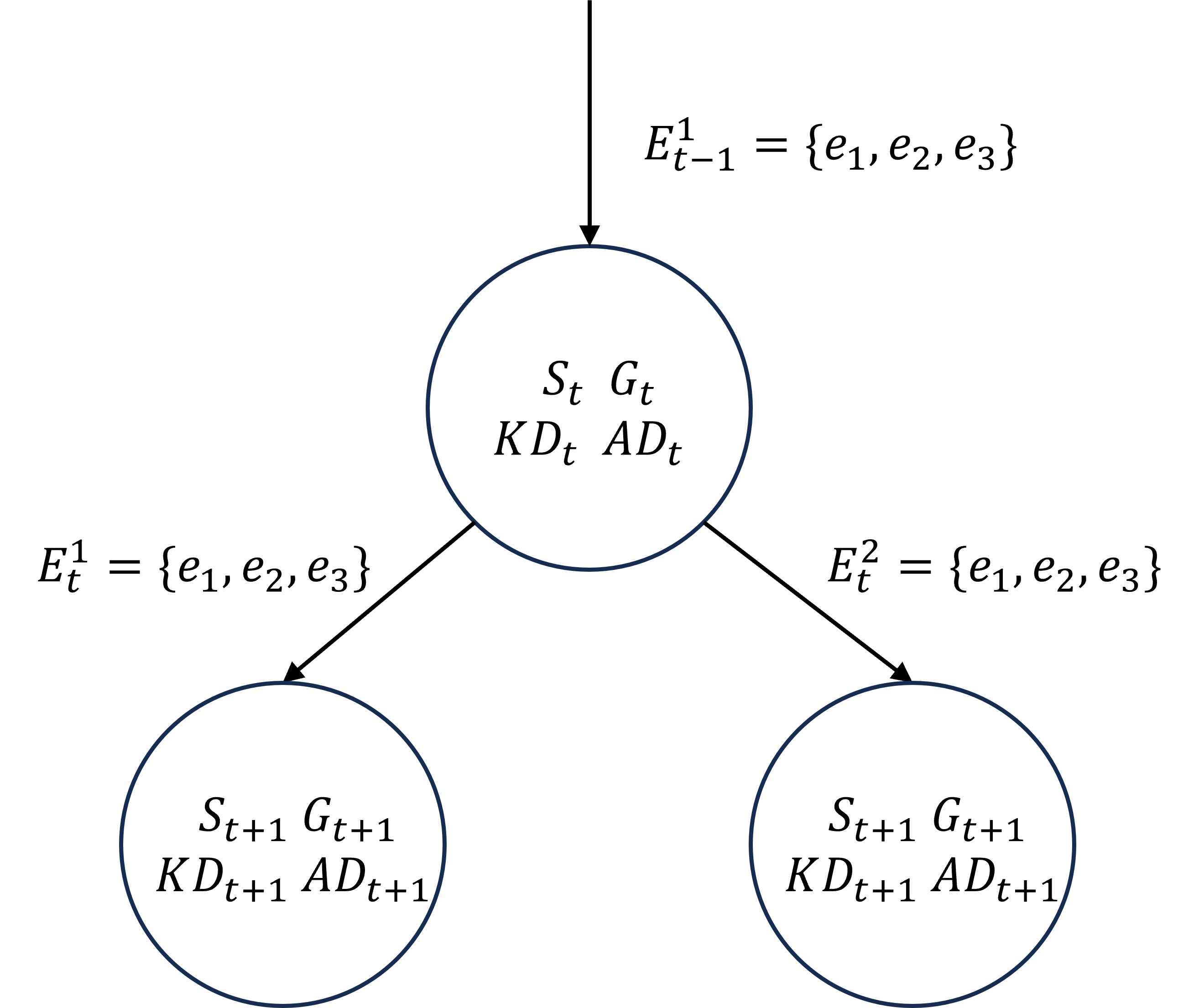}
\caption{The components of the branching plot tree. S, G, KD, and AD stand for state, goal, key decision, and alternate decision, respectively.}
\label{fig:graph}
\end{figure}

The tree (illustrated in Figure \ref{fig:graph}) is constructed as follows:
\begin{itemize}
    \item Each node contains the current \textbf{state} ($S$) and \textbf{goal} ($G$) of the character, the \textbf{key decision} ($KD$) taken in the original storyline, and an \textbf{alternate decision} ($AD$) which would lead to an alternate storyline given the same $S$ and $G$.
    \item Each edge contains a list of three sentences corresponding to events $E = \{e_1, e_2, e_3\}$ that lead the character to move from one node to another: (1) the decision made by the character---$KD$ or $AD$ as full sentences, (2) the event resulting from the decision \& leading to the next state, and (3) the next state of the character $S_{t+1}$. We will refer to these as \textbf{edge events}.
\end{itemize}

The state $S_t$ represents the current circumstances of the character after experiencing the previous events $E_{t-1}$. The goal $G_t$ is what the character tries to accomplish at the given state (defeat monster, save villagers, etc.) The key decision $KD_t$ is the decision taken in the original storyline and the alternate decision $AD_t$ is a hypothetical decision that the character could have taken to achieve $G_t$. Either decision directly leads to events $E_t$ of the corresponding edge to the next node.

In this phase (\circled{1} \& \circled{5} in Figure \ref{fig:diagram}), we prompt GPT-4 to summarize the story into a storyline in the branching plot tree and imagine the alternate decision at each node. A storyline is defined as a linear traversal of the tree from the root node to one of the leaf nodes. The strong causal relationship among these variables ensures it can capture meaningful decisions, sometimes hidden, in a plot that propels the story forward. Note that extracting events from all edges of a storyline would give us the full plot summary.

\subsection{Extracting Key Events}
\label{sec:key_event}
In our exploratory experiments, we observed that when prompted to generate alternate storylines, GPT-4 tends to imagine stories that either deviate too much from the original story or delve into specific details which leads to story stagnation. One effective solution is providing a story structure. 

A well-known story structure is the Three-Act Structure \cite{threeact} that divides the story into three phases: Setup, Confrontation, and Resolution. The backbone of this structure is three major plot points: Inciting Incident, Crisis, and Climax. These major plot points force the protagonist to experience and grow and drive the tension up and down, leading to an engaging story. 

Therefore, we prompt (Table \ref{tab:three-act}) GPT-4 to identify three key events corresponding to the three major plot points (\circled{2} in Figure \ref{fig:diagram}) from all of the edge events in a storyline and save them for later use in the prompt generation. 

\subsection{Generating Meta-Prompts}
\label{sec:prompts}
Although an ideal story should show and not tell, we observed that the LLM is prone to following instructions too explicitly. For example, the prompt ``\textit{the alternate storyline should contain new challenges...}'' would lead to events such as ``\textit{Tony faces a new challenge of...}''. Thus, we adopted a meta-prompting approach (i.e., prompting to generate prompts) which has shown to be an effective method to elicit better responses from LLMs \cite{deWynter_etal_2023,Suzgun_Kalai_2024}.

Given a storyline and every alternate decision ($AD$) in each node, we can generate prompts that are tailored to every $AD$ with concrete guiding questions using the key events we extracted from the previous phase.

Specifically, we ask GPT-4 to generate a prompt for each node, providing: (1) an alternate storyline that branches at event $e_t$ if the main character makes decision $AD$ instead of $KD$; (2) 5 concrete, guiding questions that expand: (2a) how would $AD$ change and replace the \textit{Key Events} \& (2b) how would the main character make key decisions that overcome new challenges and propel the story forward; and (3) an ideal alternate storyline. Each prompt must also: (4) strictly adhere to the given events of the original storyline; and (5) be outputted as a list of events with $AD$ as the first event.
The tables in Appendix \ref{app:example_ui} show examples of the generated prompts.

By following the guiding questions with key events from the original storyline as context, GPT-4 is able to generate coherent stories that maintain a three-act story structure, guaranteeing steady narrative pacing and character development. 

\subsection{Branching \& Merging the Tree}
\label{sec:branching}
With meta-prompts generated for each node of the original storyline, the system recursively branches out at each node and writes the alternate storylines accordingly:
\begin{enumerate}
\setlength{\itemsep}{-1pt}
    \item Extract events from all edges of the original storyline graph and include them in the prompt for the corresponding node.
    \item Generate a list of new events that continue from the previous (extracted) events. (\circled{4} in Figure \ref{fig:diagram})
    \item Create a branch from the list of new events. (\circled{5})
    \item Merge the branches of the previous events (the old plot tree) and new events to form a new branching plot tree. (\circled{6})
    \item Extract key events from the new storyline.
    \item Generate prompts for all the nodes on the branch using the procedure from Section \ref{sec:prompts}.
    \item Recursively branch out on all the nodes of the branch and write alternate storylines. Return when all meta-prompts are used.
\end{enumerate}

\begin{table}[t]
\begin{tabular}{p{3in}}
\begin{tcolorbox}
\begin{verbatim}  
'node_1': {
'state': 'Tony Stark is a wealthy 
  genius who manufactures weapons.',
'goal': “To demonstrate the new 
  Jericho missile in Afghanistan and  
  manage his company's reputation.”,
'decision': 'Tony Stark decides to go 
  to Afghanistan for the 
  demonstration.',
'edgeEvents': ['Tony Stark decides to 
  go to Afghanistan for the 
  demonstration.',
  'Stark is critically wounded and 
  captured he builds a Jericho  
  missile for them.',
  'Stark is in captivity with a 
  life-threatening injury.'],
'alternate_decision': 'Tony Stark 
  decides to send a representative to 
  demonstrate the Jericho missile, 
  while he monitors from the US.'
},
\end{verbatim}
\end{tcolorbox}
\end{tabular}
\caption{Example text generated for a node of the branching narrative based on Iron Man.}
\label{fig:node1}
\end{table}

\subsection{Narrating the Game}
\label{sec:narrating}
After our binary branching plot tree is complete, we have $2^n$ possible story endings in our game, with $n$ being the number of nodes/events in the original storyline. \whatif~traverses the tree to generate the narration for each node $n_t$ by prompting GPT-4 to narrate all events $E_{t-1}$ of the edge pointing towards the node from the previous time step. After that, it would narrate the state $S_t$ and goal $G_t$ of the character. Lastly, it provides two decisions $KD_t$ and $AD_t$ for the player to choose from. An example of this generation for a single node can be found in Table \ref{fig:node1}.  A full example can be found in Appendix \ref{app:example_ui}. 

After the player makes a decision, \whatif~continues narration using the corresponding node in the tree. Lastly, we organize all narration into an Ink file---a scripting language developed by Inkle\footnote{\url{https://www.inklestudios.com/}} for writing interactive narratives---and use InkJS\footnote{\url{https://github.com/y-lohse/inkjs}} to compile and generate the game online.

\section{Qualitative Analysis \& Discussion}
During the design of the system, we used the plot of \textit{Iron Man} taken from IMDB as the original story and compared the alternate storylines generated by GPT-4 using vanilla prompts and \whatif~prompts. 
Examples of the vanilla and \whatif~prompts and their corresponding output can be found in Appendix \ref{app:prompt_example}. We share some of our insights of this comparison below.

\subsection{Improving Branch Choices}
As the baseline, vanilla prompting converts a story plot into a list of events and uses each event as potential branching points with a simple prompt of ``\textit{Given the original storyline, imagine an alternate storyline where at event n, the main character makes decision x instead of y}'' to generate stories.
The resulting stories suffer from the following problems:

(1) \textbf{Non-choices.} Since vanilla prompting considers every event as a potential branching point, any events that do not involve decisions from the main character produce illogical branching events such as ``\textit{Stark decides to get ambushed by the Ten Rings}''. Although this aligns with the original event ``\textit{Stark is critically wounded and captured by terrorists who demand he build a Jericho missile for them.}'', this is not a decision Tony can make.

(2) \textbf{Illogical decisions.} Certain alternate decisions being proposed are impossible to consider given the circumstances and/or the character's personality. For example, after getting captured by the Ten Rings, Tony Stark would never choose to stay in captivity.

(3) \textbf{Retreat decisions.} In interactive fiction writing, there is often a retreat option where the player decides to stay out of a situation. However, this would not be an ideal branching point that is meaningful to the story as it does not move the story forward. For example, ``\textit{Tony decides not to save the villagers.}''

These decisions are attributed to branching based on events, thus the decisions are not always meaningful. \whatif~uses states and goals to capture major obstacles faced by the main character and considers the key decisions in response to the circumstances as potential branching points. This approach pushes the system to provide meaningful decisions for the player which will change the course of the story.

For example, when Tony Stark is imprisoned by Ten Rings, his state is ``\textit{in captivity with a life-threatening injury}'', and his goal is ``\textit{to survive and escape captivity}''. As an \textbf{alternate decision} is defined as a decision that the character could have made given the same state and goal, we avoid non-choices and illogical decisions by default. Since favoring a change in state forces the character to make a decision, the retreat decisions are eliminated.

\subsection{Maintaining Story Structure}
Story stagnation is the problem in fiction writing where the plot does not move forward and ends up losing the reader's interest. We observed this phenomenon in some stories from the vanilla prompt's tendency to hyper-focus on the alternate decision. For example, when Tony chooses to make a strategic plan to escape without building his famous Iron Man suit, the rest of the story turns into a journey of escaping from the cave and each event is a tiny step of the plan, with the story ending when Tony escapes and returns home.

In comparison, \whatif~uses the three-act structure to automatically extract the following three major plot points from Iron Man:
\begin{displayquote}
\textbf{Inciting Incident:} ``\textit{Tony is critically wounded in an ambush by terrorists using Stark Industries weapons}''\\
\textbf{Crisis:} ``\textit{Stane steals Stark's arc reactor, leaving him to die.}''\\
\textbf{Climax: }``\textit{Stark survives using his original reactor, and battles Stane at Stark Industries}''
\end{displayquote}

These three key events are essential in creating the story arc of the original story, and the alternate stories should have similar events that build up tension, challenge the main character, and eventually come to a resolution. We include these events in the meta-prompt generation phase and ask GPT-4 to provide concrete guiding questions that expand on how the generated alternate decision is changed or replaced. With this prompt, for the alternate decision of ``\textit{building the missile for Ten Rings hoping for escape later}'', GPT-4 generated:
\begin{displayquote}
1. How would this alternate decision impact Stark's relationship with Yinsen and their plan for escape? \\
2. What consequences would arise from the terrorists obtaining a functioning Jericho missile?\\
3. How would Stark manage to escape without the assistance of a suit of armor?\\
4. In what ways would Stark confront and expose Stane's dealings without the identity of Iron Man?\\
5. How would Stark's decision affect his ideology and future decisions regarding weapon manufacturing and his role in global conflicts?
\end{displayquote}

By answering these concrete guiding questions specifically related to major plot points from the original story, GPT-4 ensures that (1) the consequence of the alternate questions are extensively explored, (2) the unaffected original events still occur, e.g. Stane's betrayal, and (3) the changes to the story arc are considered.

After new events are generated, they are organized into a branching plot tree again. A new set of key events is extracted from the merged tree and fed to generate more context-specific prompts. We make sure to only include the key events that have not yet occurred at the branching event to prevent changing the past. 

\subsection{Controllable LLM outputs}
While text-adventure games where any action could potentially cause the player's demise, many modern interactive fictions prefer to balance branching narratives, where each branch has approximately the same length to prevent the story from ending too quickly. \whatif~builds a full binary branching plot tree to ensure complete balance. Furthermore, GPT-4 is told to keep the generated alternate plots to the same number of events as the original plot. To control GPT-4's output, we use the JSON mode with predefined JSON schema for all prompting functions to ensure the correct number of events or nodes are generated. Examples of the JSON schema are included in Appendix \ref{app:json}.

\section{Conclusion}
In this paper, we presented an interactive fiction system called \whatif, which uses LLMs to generate alternate timelines from pre-written storylines. \whatif~focuses on character-driven generation to create engaging branching points in the narratives, uses meta-prompting to consider the narrative from a broader perspective using the three-act story structure, and maintains consistent storytelling using the meta-prompts and an external graph structure. We hope to see future story generation systems use meta-prompting to encourage controlled, long-form generation.

\section*{Limitations}
Two of the main limitations of using our system are time and money. Branching takes about a minute to generate each branch. We opted to generate the whole tree ahead of time---which saves time while playing---at the expense of waiting for longer to start.

GPT-4 is priced by how many tokens it has used as input and how many tokens it has generated. This makes any widespread use of a system that relies on GPT-4 to be impractical. If we were to release \whatif~to the general public, it would need to be converted to a free-to-use LLM of similar ability. \whatif~might need to be converted to a few-shot system, but the overall structure and prompting techniques we used will most likely still help the LLM.

Also, as with all large language models, the generated text cannot be predicted ahead of time, which can lead to potentially harmful stories.

Lastly, our experiments with \whatif~were in English only. Because of its architecture, \whatif~is bound to the LLM underneath it. As most LLMs work best with English at the moment, this limits our system as well.


\bibliography{anthology,custom}

\begin{thebibliography}{51}
\expandafter\ifx\csname natexlab\endcsname\relax\def\natexlab#1{#1}\fi

\bibitem[{Ammanabrolu et~al.(2020)Ammanabrolu, Cheung, Tu, Broniec, and Riedl}]{Ammanabrolu_etal_2020_bringing}
Prithviraj Ammanabrolu, Wesley Cheung, Dan Tu, William Broniec, and Mark~O. Riedl. 2020.
\newblock \href {https://doi.org/10.1609/aiide.v16i1.7400} {{Bringing Stories Alive}: Generating interactive fiction worlds}.
\newblock \emph{AAAI Conference on Artificial Intelligence and Interactive Digital Entertainment (AIIDE)}, 16(1):3–9.

\bibitem[{Ammanabrolu et~al.(2021)Ammanabrolu, Urbanek, Li, Szlam, Rocktäschel, and Weston}]{Ammanabrolu_etal_2021}
Prithviraj Ammanabrolu, Jack Urbanek, Margaret Li, Arthur Szlam, Tim Rocktäschel, and Jason Weston. 2021.
\newblock \href {https://doi.org/10.18653/v1/2021.naacl-main.64} {{How to Motivate Your Dragon}: Teaching goal-driven agents to speak and act in fantasy worlds}.
\newblock In \emph{The North American Chapter of the Association for Computational Linguistics (NAACL-HLT)}, page 807–833, Virtual Event. Association for Computational Linguistics.

\bibitem[{Ashby et~al.(2023)Ashby, Webb, Knapp, Searle, and Fulda}]{Ashby_Webb_Knapp_Searle_Fulda_2023}
Trevor Ashby, Braden Webb, Gregory Knapp, Jackson Searle, and Nancy Fulda. 2023.
\newblock \href {https://doi.org/10.1145/3544548.3581441} {Personalized quest and dialogue generation in role-playing games: A knowledge graph- and language model-based approach}.
\newblock In \emph{ACM Conference on Human Factors in Computing Systems (CHI)}, pages 290:1--290:20.

\bibitem[{Callison-Burch et~al.(2022)Callison-Burch, Singh~Tomar, Martin, Ippolito, Bailis, and Reitter}]{ccb2022dungeons}
Chris Callison-Burch, Gaurav Singh~Tomar, Lara~J. Martin, Daphne Ippolito, Suma Bailis, and David Reitter. 2022.
\newblock \href {https://doi.org/10.18653/v1/2022.emnlp-main.637} {{Dungeons and Dragons} as a dialogue challenge for artificial intelligence}.
\newblock In \emph{{Conference on Empirical Methods in Natural Language Processing (EMNLP)}}, pages 9379--9393, Abu Dhabi, UAE. ACL.

\bibitem[{De~Lima et~al.(2021)De~Lima, Feijo, and Furtado}]{DeLima_Feijo_Furtado_2021}
Edirlei~Soares De~Lima, Bruno Feijo, and Antonio~L. Furtado. 2021.
\newblock \href {https://doi.org/10.1109/SBGames54170.2021.00012} {Adaptive branching quests based on automated planning and story arcs}.
\newblock In \emph{Brazilian Symposium on Computer Games and Digital Entertainment (SBGames)}, pages 9--18, Gramado, Brazil. IEEE.

\bibitem[{de~Wynter et~al.(2023)de~Wynter, Wang, Gu, and Chen}]{deWynter_etal_2023}
Adrian de~Wynter, Xun Wang, Qilong Gu, and Si-Qing Chen. 2023.
\newblock \href {https://doi.org/10.48550/arXiv.2312.06562} {On meta-prompting}.
\newblock \emph{arXiv preprint arXiv:2312.06562}.

\bibitem[{Dong et~al.(2023)Dong, Martin, and Callison-Burch}]{Dong_etal_2023}
Yijiang~River Dong, Lara~J. Martin, and Chris Callison-Burch. 2023.
\newblock \href {https://doi.org/10.18653/v1/2023.findings-acl.832} {{CoRRPUS}: Code-based structured prompting for neurosymbolic story understanding}.
\newblock In \emph{Findings of the Association for Computational Linguistics: ACL 2023}, pages 13152--13168, Toronto, Canada. ACL.

\bibitem[{Farrell and Ware(2024)}]{Farrell_2024}
Rachelyn Farrell and Stephen~G. Ware. 2024.
\newblock \href {https://doi.org/10.36227/techrxiv.171085113.35202301/v1} {{Planning Stories Neurally}}.

\bibitem[{Field(1979)}]{threeact}
Syd Field. 1979.
\newblock \emph{Screenplay: The Foundations of Screenwriting}.
\newblock Dell Publishing Company.

\bibitem[{Gero et~al.(2023)Gero, Long, and Chilton}]{Gero_Long_Chilton_2023}
Katy~Ilonka Gero, Tao Long, and Lydia Chilton. 2023.
\newblock \href {https://doi.org/10.1145/3544548.3580782} {Social dynamics of {AI} support in creative writing}.
\newblock In \emph{ACM Conference on Human Factors in Computing Systems (CHI)}, pages 245:1--245:15, Hamburg, Germany. ACM.

\bibitem[{Harmon and Rutman(2023)}]{Harmon_Rutman_2023}
Sarah Harmon and Sophia Rutman. 2023.
\newblock \href {https://doi.org/10.1007/978-3-031-47655-6_13} {Prompt engineering for narrative choice generation}.
\newblock In \emph{International Conference on Interactive Digital Storytelling (ICIDS)}, volume 14383 of \emph{Lecture Notes in Computer Science}, pages 208--225, Kobe, Japan. Springer Nature Switzerland.

\bibitem[{Johnson-Bey et~al.(2023)Johnson-Bey, Mateas, and Wardrip-Fruin}]{Johnson-Bey_etal_2023}
Shi Johnson-Bey, Michael Mateas, and Noah Wardrip-Fruin. 2023.
\newblock \href {https://www.exag.org/papers/Toward%20Using%20ChatGPT%20to%20Generate%20Theme-Relevant%20Simulated%20Storyworlds.pdf} {Toward using {ChatGPT} to generate theme-relevant simulated storyworlds}.
\newblock In \emph{AIIDE Workshop on Experimental Artificial Intelligence in Games (EXAG)}, Salt Lake City, UT, USA.

\bibitem[{Kelly et~al.(2023)Kelly, Calderwood, Wardrip-Fruin, and Mateas}]{Kelly_etal_2023}
Jack Kelly, Alex Calderwood, Noah Wardrip-Fruin, and Michael Mateas. 2023.
\newblock \href {https://doi.org/10.1609/aiide.v19i1.27502} {{There and Back Again}: Extracting formal domains for controllable neurosymbolic story authoring}.
\newblock \emph{AAAI Conference on Artificial Intelligence and Interactive Digital Entertainment (AIIDE)}, 19(11):64--74.

\bibitem[{Kumaran et~al.(2024)Kumaran, Rowe, and Lester}]{Kumaran_2024}
Vikram Kumaran, Jonathan Rowe, and James Lester. 2024.
\newblock \href {https://doi.org/10.1609/aiide.v20i1.31868} {{NarrativeGenie: Generating Narrative Beats and Dynamic Storytelling with Large Language Models}}.
\newblock \emph{Proceedings of the AAAI Conference on Artificial Intelligence and Interactive Digital Entertainment}, 20(11):76–86.

\bibitem[{Kumaran et~al.(2023)Kumaran, Rowe, Mott, and Lester}]{Kumaran_etal_2023}
Vikram Kumaran, Jonathan Rowe, Bradford Mott, and James Lester. 2023.
\newblock \href {https://doi.org/10.1609/aiide.v19i1.27504} {{SceneCraft}: Automating interactive narrative scene generation in digital games with large language models}.
\newblock \emph{AAAI Conference on Artificial Intelligence and Interactive Digital Entertainment (AIIDE)}, 19(11):86–96.

\bibitem[{Lal et~al.(2021)Lal, Chambers, Mooney, and Balasubramanian}]{Lal_Chambers_Mooney_Balasubramanian_2021}
Yash~Kumar Lal, Nathanael Chambers, Raymond Mooney, and Niranjan Balasubramanian. 2021.
\newblock \href {https://aclanthology.org/2021.findings-acl.53} {{TellMeWhy}: A dataset for answering why-questions in narratives}.
\newblock In \emph{Findings of the Association for Computational Linguistics: ACL-IJCNLP}, page 596–610, Online. ACL.

\bibitem[{Lebowitz(1984)}]{Lebowitz_1984}
Michael Lebowitz. 1984.
\newblock \href {https://doi.org/10.1016/0304-422X(84)90001-9} {Creating characters in a story-telling universe}.
\newblock \emph{Poetics}, 13(3):171–194.

\bibitem[{Louis and Sutton(2018)}]{Louis_Sutton_2018}
Annie Louis and Charles Sutton. 2018.
\newblock \href {https://doi.org/10.18653/v1/N18-2111} {{Deep Dungeons and Dragons}: Learning character-action interactions from role-playing game transcripts}.
\newblock In \emph{Conference of the North American Chapter of the Association for Computational Linguistics: Human Language Technologies (NAACL)}, volume Volume 2 (Short Papers), page 708–713, New Orleans, Louisiana. ACL.

\bibitem[{Martin(2021)}]{Martin_2021}
Lara~J. Martin. 2021.
\newblock \href {https://smartech.gatech.edu/handle/1853/64643} {\emph{Neurosymbolic Automated Story Generation}}.
\newblock Phd, Georgia Institute of Technology, Atlanta, GA.

\bibitem[{Martin et~al.(2018)Martin, Sood, and Riedl}]{Martin_Sood_Riedl_2018}
Lara~J. Martin, Srijan Sood, and Mark~O. Riedl. 2018.
\newblock \href {http://ceur-ws.org/Vol-2321/paper4.pdf} {{Dungeons and DQNs}: Toward reinforcement learning agents that play tabletop roleplaying games}.
\newblock In \emph{Joint Workshop on Intelligent Narrative Technologies and Workshop on Intelligent Cinematography and Editing (INT-WICED)}, Edmonton, AB, Canada. http://ceur-ws.org.

\bibitem[{Meehan(1977)}]{Meehan_1977}
James~R. Meehan. 1977.
\newblock \href {https://dl.acm.org/doi/10.5555/1624435.1624452} {{TALE-SPIN}, an interactive program that writes stories}.
\newblock In \emph{International Joint Conference on Artificial Intelligence (IJCAI)}, volume~1, page 91–98. ACM.

\bibitem[{Méndez and Gervás(2023)}]{Méndez_Gervás_2023}
Gonzalo Méndez and Pablo Gervás. 2023.
\newblock \href {https://computationalcreativity.net/iccc23/papers/ICCC-2023_paper_124.pdf} {Using {ChatGPT} for story sifting in narrative generation}.
\newblock In \emph{International Conference on Computational Creativity (ICCC)}, Ontario, Canada.

\bibitem[{Park et~al.(2023)Park, Shin, Kim, and Bae}]{Park_Shin_Kim_Bae_2023}
Jeongyoon Park, Jumin Shin, Gayeon Kim, and Byung-Chull Bae. 2023.
\newblock \href {https://doi.org/10.1007/978-3-031-47658-7_22} {Designing a language model-based authoring tool prototype for interactive storytelling}.
\newblock In \emph{International Conference on Interactive Digital Storytelling (ICIDS)}, volume 14384 of \emph{Lecture Notes in Computer Science}, page 239–245, Kobe, Japan. Springer Nature Switzerland.

\bibitem[{Pei et~al.(2024)Pei, Patel, El-Refai, and Li}]{Patel_etal_2024}
Jonathan Pei, Zeeshan Patel, Karim El-Refai, and Tianle Li. 2024.
\newblock \href {https://doi.org/10.18653/v1/2024.findings-emnlp.824} {{SWAG}: Storytelling with action guidance}.
\newblock In \emph{Findings of the Association for Computational Linguistics: EMNLP 2024}, pages 14086--14106, Miami, Florida, USA. Association for Computational Linguistics.

\bibitem[{Qin and Zhao(2022)}]{Qin_Zhao_2022}
Wentao Qin and Dongyan Zhao. 2022.
\newblock \href {https://doi.org/10.1007/978-3-031-17120-8_28} {{Retrieval, Selection and Writing}: A three-stage knowledge grounded storytelling model}.
\newblock In \emph{Natural Language Processing and Chinese Computing}, Lecture Notes in Computer Science, page 352–363. Springer International Publishing.

\bibitem[{Rafailov et~al.(2023)Rafailov, Sharma, Mitchell, Manning, Ermon, and Finn}]{DPO}
Rafael Rafailov, Archit Sharma, Eric Mitchell, Christopher~D. Manning, Stefano Ermon, and Chelsea Finn. 2023.
\newblock \href {https://proceedings.neurips.cc/paper_files/paper/2023/hash/a85b405ed65c6477a4fe8302b5e06ce7-Abstract-Conference.html} {Direct preference optimization: Your language model is secretly a reward model}.
\newblock In \emph{Advances in Neural Information Processing Systems (NeurIPS)}, volume~36, page 53728–53741, New Orleans, LA, USA. Curran Associates, Inc.

\bibitem[{Razumovskaia et~al.(2024)Razumovskaia, Maynez, Louis, Lapata, and Narayan}]{razumovskaia-etal-2024-little}
Evgeniia Razumovskaia, Joshua Maynez, Annie Louis, Mirella Lapata, and Shashi Narayan. 2024.
\newblock \href {https://aclanthology.org/2024.lrec-main.929} {Little red riding hood goes around the globe: Crosslingual story planning and generation with large language models}.
\newblock In \emph{Proceedings of the 2024 Joint International Conference on Computational Linguistics, Language Resources and Evaluation (LREC-COLING 2024)}, pages 10616--10631, Torino, Italia. ELRA and ICCL.

\bibitem[{Riedl and Bulitko(2013)}]{Riedl_Bulitko_2013}
Mark~O. Riedl and Vadim Bulitko. 2013.
\newblock \href {https://doi.org/10.1609/aimag.v34i1.2449} {Interactive narrative: An intelligent systems approach}.
\newblock \emph{AI Magazine}, 34(1):67--77.

\bibitem[{Sanghrajka(2019)}]{Sanghrajka_2019}
Rushit Sanghrajka. 2019.
\newblock \href {https://doi.org/10.1609/aiide.v15i1.5251} {Interactive narrative authoring using cognitive models in narrative planning}.
\newblock \emph{AAAI Conference on Artificial Intelligence and Interactive Digital Entertainment (AIIDE): Doctoral Consortium Abstracts}, 15(1):224--226.

\bibitem[{See et~al.(2019)See, Pappu, Saxena, Yerukola, and Manning}]{See_etal_2019}
Abigail See, Aneesh Pappu, Rohun Saxena, Akhila Yerukola, and Christopher~D. Manning. 2019.
\newblock \href {https://doi.org/10.18653/v1/K19-1079} {Do massively pretrained language models make better storytellers?}
\newblock In \emph{Conference on Computational Natural Language Learning (CoNLL)}, page 843–861, Hong Kong, China. ACL.

\bibitem[{Shao et~al.(2023)Shao, Li, Dai, and Qiu}]{Shao_Li_Dai_Qiu_2023}
Yunfan Shao, Linyang Li, Junqi Dai, and Xipeng Qiu. 2023.
\newblock \href {https://doi.org/10.48550/arXiv.2310.10158} {{Character-LLM}: A trainable agent for role-playing}.
\newblock In \emph{Conference on Empirical Methods in Natural Language Processing (EMNLP)}, page 13153–13187, Singapore. ACL.

\bibitem[{Simon and Muise(2022)}]{Simon_Muise_2022}
Nisha Simon and Christian Muise. 2022.
\newblock \href {https://icaps22.icaps-conference.org/workshops/SPARK/papers/spark2022_paper_2.pdf} {{TattleTale}: Storytelling with planning and large language models}.
\newblock In \emph{Workshop on Scheduling and Planning Applications woRKshop (SPARK) at ICAPS}, Virtual.

\bibitem[{Suzgun and Kalai(2024)}]{Suzgun_Kalai_2024}
Mirac Suzgun and Adam~Tauman Kalai. 2024.
\newblock \href {https://doi.org/10.48550/arXiv.2401.12954} {{Meta-Prompting}: Enhancing language models with task-agnostic scaffolding}.
\newblock \emph{arXiv preprint arXiv:2401.12954}.

\bibitem[{Swanson et~al.(2021)Swanson, Mathewson, Pietrzak, Chen, and Dinalescu}]{Swanson_etal_2021}
Ben Swanson, Kory~W. Mathewson, Ben Pietrzak, Sherol Chen, and Monica Dinalescu. 2021.
\newblock \href {https://doi.org/10.18653/v1/2021.eacl-demos.29} {{Story Centaur}: Large language model few shot learning as a creative writing tool}.
\newblock In \emph{Conference of the European Chapter of the Association for Computational Linguistics (EACL): System Demonstrations}, page 244–256, Online. ACL.

\bibitem[{Walton(2019)}]{AIDungeon}
Nick Walton. 2019.
\newblock \href {https://play.aidungeon.com/} {{AI Dungeon}}.

\bibitem[{Wang et~al.(2024)Wang, Peng, Que, Liu, Zhou, Wu, Guo, Gan, Ni, Yang, Zhang, Zhang, Ouyang, Xu, Huang, Fu, and Peng}]{Wang_etal_2023}
Noah Wang, Z.y. Peng, Haoran Que, Jiaheng Liu, Wangchunshu Zhou, Yuhan Wu, Hongcheng Guo, Ruitong Gan, Zehao Ni, Jian Yang, Man Zhang, Zhaoxiang Zhang, Wanli Ouyang, Ke~Xu, Wenhao Huang, Jie Fu, and Junran Peng. 2024.
\newblock \href {https://doi.org/10.18653/v1/2024.findings-acl.878} {{R}ole{LLM}: Benchmarking, eliciting, and enhancing role-playing abilities of large language models}.
\newblock In \emph{Findings of the Association for Computational Linguistics: ACL 2024}, pages 14743--14777, Bangkok, Thailand. Association for Computational Linguistics.

\bibitem[{Wang et~al.(2025)Wang, Hu, Li, Wang, Li, Hu, and Tan}]{Wang_2025}
Qianyue Wang, Jinwu Hu, Zhengping Li, Yufeng Wang, Daiyuan Li, Yu~Hu, and Mingkui Tan. 2025.
\newblock \href {https://aclanthology.org/2025.naacl-long.63/} {{Generating Long-form Story Using Dynamic Hierarchical Outlining with Memory-Enhancement}}.
\newblock In \emph{Proceedings of the 2025 Conference of the Nations of the Americas Chapter of the Association for Computational Linguistics: Human Language Technologies}, volume Volume 1: Long Papers, page 1352–1391, Albuquerque, New Mexico. Association for Computational Linguistics.

\bibitem[{Ware and Siler(2021)}]{Ware_Siler_2021}
Stephen~G. Ware and Cory Siler. 2021.
\newblock \href {https://doi.org/10.1609/aiide.v17i1.18896} {Sabre: A narrative planner supporting intention and deep theory of mind}.
\newblock \emph{AAAI Conference on Artificial Intelligence and Interactive Digital Entertainment (AIIDE)}, 17(1):99--106.

\bibitem[{Wen et~al.(2023)Wen, Tian, Wu, Yang, Shi, Huang, and Li}]{wen-etal-2023-grove}
Zhihua Wen, Zhiliang Tian, Wei Wu, Yuxin Yang, Yanqi Shi, Zhen Huang, and Dongsheng Li. 2023.
\newblock \href {https://doi.org/10.18653/v1/2023.findings-emnlp.262} {{GROVE}: A retrieval-augmented complex story generation framework with a forest of evidence}.
\newblock In \emph{Findings of the Association for Computational Linguistics: EMNLP 2023}, pages 3980--3998, Singapore. Association for Computational Linguistics.

\bibitem[{Xia et~al.(2025)Xia, Peng, Qi, Wang, Xu, Hou, and Li}]{Xia_2025}
Haotian Xia, Hao Peng, Yunjia Qi, Xiaozhi Wang, Bin Xu, Lei Hou, and Juanzi Li. 2025.
\newblock \href {https://doi.org/10.48550/arXiv.2506.16445} {{StoryWriter: A Multi-Agent Framework for Long Story Generation}}.
\newblock \emph{arXiv preprint arXiv:2506.16445}.

\bibitem[{Xie et~al.(2023)Xie, Cohn, and Lau}]{xie-etal-2023-next}
Zhuohan Xie, Trevor Cohn, and Jey~Han Lau. 2023.
\newblock \href {https://doi.org/10.18653/v1/2023.inlg-main.23} {The next chapter: A study of large language models in storytelling}.
\newblock In \emph{Proceedings of the 16th International Natural Language Generation Conference}, pages 323--351, Prague, Czechia. Association for Computational Linguistics.

\bibitem[{Yan et~al.(2023)Yan, Li, Zhang, Wang, Yang, and Yan}]{Yan_Li_Zhang_Wang_Yang_Yan_2023}
Ming Yan, Ruihao Li, Hao Zhang, Hao Wang, Zhilan Yang, and Ji~Yan. 2023.
\newblock \href {https://doi.org/10.48550/arXiv.2312.17653} {{LARP}: Language-agent role play for open-world games}.
\newblock \emph{arXiv preprint arXiv:2312.17653}.

\bibitem[{Yang et~al.(2022)Yang, Tian, Peng, and Klein}]{Yang_etal_2022}
Kevin Yang, Yuandong Tian, Nanyun Peng, and Dan Klein. 2022.
\newblock \href {https://doi.org/10.18653/v1/2022.emnlp-main.296} {{Re$^3$}: Generating longer stories with recursive reprompting and revision}.
\newblock In \emph{Conference on Empirical Methods in Natural Language Processing (EMNLP)}, page 4393–4479, Abu Dhabi, United Arab Emirates. ACL.

\bibitem[{Ye et~al.(2023)Ye, Cui, Shi, and Riedl}]{Ye_etal_2023}
Anbang Ye, Christopher Cui, Taiwei Shi, and Mark~O. Riedl. 2023.
\newblock \href {https://doi.org/10.48550/arXiv.2212.08718} {Neural story planning}.
\newblock In \emph{AAAI Workshop on Creative AI Across Modalities}. AAAI.

\bibitem[{Young et~al.(2013)Young, Ware, Cassell, and Robertson}]{Young_etal_2013}
R.~Michael Young, Stephen~G. Ware, Bradley~A. Cassell, and Justus Robertson. 2013.
\newblock Plans and planning in narrative generation: a review of plan-based approaches to the generation of story, discourse and interactivity in narratives.
\newblock \emph{Sprache und Datenverarbeitung, Special Issue on Formal and Computational Models of Narrative}, 37(1–2):41–64.

\bibitem[{Yu et~al.(2025)Yu, Shi, Zhao, and Penn}]{Yu_2025}
Tian Yu, Ken Shi, Zixin Zhao, and Gerald Penn. 2025.
\newblock \href {https://aclanthology.org/2025.in2writing-1.9/} {{Multi-Agent Based Character Simulation for Story Writing}}.
\newblock In \emph{Proceedings of the Fourth Workshop on Intelligent and Interactive Writing Assistants (In2Writing 2025)}, page 87–108, Albuquerque, New Mexico, US. Association for Computational Linguistics.

\bibitem[{Yuan et~al.(2022)Yuan, Coenen, Reif, and Ippolito}]{Yuan_etal_2022_Wordcraft}
Ann Yuan, Andy Coenen, Emily Reif, and Daphne Ippolito. 2022.
\newblock \href {https://doi.org/10.1145/3490099.3511105} {Wordcraft: Story writing with large language models}.
\newblock In \emph{International Conference on Intelligent User Interfaces (IUI)}, page 841–852, Helsinki, Finland. ACL.

\bibitem[{Zhou et~al.(2025)Zhou, Basavatia, Siam, Chen, and Riedl}]{Zhou_2025}
Eric Zhou, Shreyas Basavatia, Moontashir Siam, Zexin Chen, and Mark~O. Riedl. 2025.
\newblock \href {https://doi.org/10.48550/arXiv.2505.03547} {{STORY2GAME: Generating (Almost) Everything in an Interactive Fiction Game}}.
\newblock \emph{arXiv preprint arXiv:2505.03547}.

\bibitem[{Zhu et~al.(2023{\natexlab{a}})Zhu, Martin, Head, and Callison-Burch}]{CALYPSO}
Andrew Zhu, Lara~J. Martin, Andrew Head, and Chris Callison-Burch. 2023{\natexlab{a}}.
\newblock \href {https://doi.org/10.1609/aiide.v19i1.27534} {{CALYPSO}: {LLMs} as dungeon masters’ assistants}.
\newblock \emph{AAAI Conference on Artificial Intelligence and Interactive Digital Entertainment (AIIDE)}, 19(1):380–390.

\bibitem[{Zhu et~al.(2023{\natexlab{b}})Zhu, Cohen, Wang, Yang, Yang, Jiao, and Tian}]{Zhu_etal_2023_End}
Hanlin Zhu, Andrew Cohen, Danqing Wang, Kevin Yang, Xiaomeng Yang, Jiantao Jiao, and Yuandong Tian. 2023{\natexlab{b}}.
\newblock \href {https://doi.org/10.48550/arXiv.2310.08796} {End-to-end story plot generator}.
\newblock \emph{arXiv preprint arXiv:2310.08796}.

\bibitem[{Zimmerman et~al.(2025)Zimmerman, Sahu, and Vechtomova}]{Zimmerman_2025}
Brian Zimmerman, Gaurav Sahu, and Olga Vechtomova. 2025.
\newblock \href {https://doi.org/10.1007/978-3-031-90167-6_16} {{Future Sight: Fine-Tuning Language Models for Dynamic Story Generation}}.
\newblock In \emph{Artificial Intelligence in Music, Sound, Art and Design}, page 233–248, Trieste, Italy. Springer Nature Switzerland.

\end{thebibliography}

\appendix

\begin{table*}
\begin{minipage}{\textwidth}
\section{Example Generation}
\label{app:example_ui}
The following is the JSON generated by \whatif~using the prompt from Figure \ref{fig:UI} (plot from Iron Man). They are the extracted key events (Table \ref{tab:key_events}), the generated meta-prompt (Table \ref{tab:metaprompt}), the new storyline (Table \ref{tab:new_story}), and the final game narration (Table \ref{tab:narration}), respectively.

\begin{tabular}{p{6.25in}}
\begin{tcolorbox}
\begin{verbatim}  
'node_1': {
       'state': 'Tony Stark is a wealthy genius who manufactures weapons.',
       'goal': “To demonstrate the new Jericho missile in Afghanistan and  
          manage his company's reputation.”,
       'decision': 'Tony Stark decides to go to Afghanistan for the 
          demonstration.',
       'edgeEvents': ['Tony Stark decides to go to Afghanistan for the 
          demonstration.',
          'Stark is critically wounded and captured by terrorists who demand 
          he build a Jericho missile for them.',
          'Stark is in captivity with a life-threatening injury.'],
       'alternate_decision': 'Tony Stark decides to send a representative to 
          demonstrate the Jericho missile, while he monitors from the US.'
      },
  
'node_2': {
       'state': 'Stark is in captivity with a life-threatening injury.',
       'goal': 'To survive and escape captivity without building the missile   
          for the terrorists.',
       'decision': 'Tony Stark decides to build an armored suit instead of  
          the missile.',
       'edgeEvents': ['Tony Stark decides to build an armored suit instead  
          of the missile.',
          'Stark and Yinsen work on the suit secretly, using available
          resources.',
          'Stark escapes using the suit but Yinsen dies.'],
       'alternate_decision': 'Tony Stark decides to build the missile
          as requested, planning to escape afterward.'
      },
  
'node_3': {
       'state': 'Stark escapes using the suit but Yinsen dies.',
       'goal': 'To stop Stark Industries from manufacturing weapons   
          that end up in the wrong hands.',
       'decision': 'Tony Stark decides to stop manufacturing weapons  
          at Stark Industries.',
       'edgeEvents': ['Tony Stark decides to stop manufacturing weapons at  
          Stark Industries.',
          'Stark returns home and announces the cessation of weapon 
          manufacturing.',
          “Obadiah Stane opposes Stark's new directive.”],
       'alternate_decision': 'Tony Stark decides to increase security 
           and oversight of weapon sales instead of halting production.'
      },
\end{verbatim}
\end{tcolorbox}
\end{tabular}
\end{minipage}
\end{table*}

\begin{table*}

    \begin{tabular}{p{6.25in}}
    \begin{tcolorbox}
         \begin{verbatim}
'node_4': {
       'state': “Obadiah Stane opposes Stark's new directive.”,
       'goal': “To find evidence of Stane's betrayal and stop him.”,
       'decision': “Tony Stark decides to improve his suit and find proof of 
           Stane's treachery.”,
        'edgeEvents': [“Tony Stark decides to improve his suit and find proof of 
           Stane's treachery.”,
           “Stark saves Yinsen's village and discovers Stane's involvement  
           with the terrorists.”,
           “Stark realizes Stane's intentions and confronts him.”],
       'alternate_decision': 'Tony Stark decides to confront Stane directly
           about his suspicions without gathering evidence.'
      },
         
'node_5': {
       'state': “Stark realizes Stane's intentions and confronts him.”,
       'goal': 'To stop Stane from using the technology for evil and 
          protect his legacy.',
       'decision': 'Tony Stark decides to fight Stane, who has built his own 
          suit.',
       'edgeEvents': ['Tony Stark decides to fight Stane, who has built his own
          suit.',
          'During their confrontation, Stark instructs Potts to overload the  
          arc reactor.',
          “Stane is defeated but Stark's identity as Iron Man is revealed.”],
       'alternate_decision': 'Tony Stark decides to seek the help of authorities  
          and expose Stane publicly without a direct confrontation.'
      },
         
'node_6': {
       'state': “Stane is defeated but Stark's identity as Iron Man is 
          revealed.”,
       'goal': 'To manage the fallout of his identity as Iron Man being 
          revealed.',
       'decision': 'Tony Stark decides to publicly admit he is Iron Man.',
       'edgeEvents': ['Tony Stark decides to publicly admit he is Iron Man.',
          'At a press conference, Stark announces his identity, contrasting 
          the offered cover story.',
          'Stark adjusts to his new role as a publicly known superhero.'],
       'alternate_decision': 'Tony Stark decides to stick to the cover story  
          provided and hide his identity as Iron Man.'
      }
\end{verbatim}
\end{tcolorbox}   
         \caption{The JSON of key events generated from the plot Iron Man taken from Wikipedia. See Section \ref{sec:key_event} for how key events are extracted.}
         \label{tab:key_events}
    \end{tabular}   
 
\end{table*}

\begin{table*}

    \begin{tabular}{p{6.25in}}
    \begin{tcolorbox}
         \begin{verbatim}   
Using the original storyline as a reference, write an alternate storyline that 
branches out at event 4 with Tony Stark deciding to build the missile as 
requested, planning to escape afterward, instead of deciding to build an armored 
suit. As you craft this new narrative, consider and incorporate answers to the  
following thought-provoking questions: 

1. How would building the missile instead of the armored suit impact Stark's  
inventive strategy for escape and survival? 
2. In what ways could this decision affect Stark's moral journey and his  
relationship with Yinsen? 
3. Would Stark still have the motivation to stop manufacturing weapons upon his  
return, and how would this decision manifest? 
4. How would Stark's eventual discovery of Stane's betrayal and eventual  
confrontation play out differently? 
5. How might Stark approach his role as a superhero and his public persona in 
this alternate scenario? 

Describe what an ideal alternate storyline should look like, encapsulating 
Stark's complex character development, thrilling escape, consequential actions   
upon returning home, and a nuanced showdown with Stane. Then, output    
the alternate storyline as a list of 15 events, starting with Tony Stark decides   
to build the missile as requested, planning to escape afterward. Ensure the   
storyline comprehensively covers new challenges Stark faces, key decisions he   
makes to overcome these challenges, and how these contribute to moving the story
forward.
\end{verbatim}
\end{tcolorbox}  
\caption{The generated meta-prompt for altering the timeline of Iron Man. See Section \ref{sec:prompts} for how the meta-prompt is generated.}
\label{tab:metaprompt}
    \end{tabular}     
\end{table*}


\begin{table*}
\begin{tabular}{p{6.25in}}
\begin{tcolorbox}
\begin{verbatim}
[
'Tony Stark decides to go to Afghanistan for the demonstration.',
'Stark is critically wounded and captured by terrorists who demand he build a 
Jericho missile for them.',
'Stark is in captivity with a life-threatening injury.',
'Tony Stark decides to build the missile as requested, planning to escape 
afterward.',
“Stark secretly designs a fail-safe within the missile to ensure it won't 
detonate.”,
'During the missile building, Stark and Yinsen form a deep bond, discussing the 
impact of weapons on the world.',
'Yinsen shares his technical knowledge, aiding Stark in creating a concealed, 
miniaturized drone for their escape.',
'They initiate their escape using the drone to create a diversion. Yinsen 
sacrifices himself in a moment of distraction for Stark to flee.',
“Haunted by Yinsen's sacrifice, Stark is rescued and returns home with a heavy 
conscience.”,
'Stark decides to halt weapon production at Stark Industries, facing internal 
opposition.',
“Discovering Stane's involvement with terrorists, Stark is conflicted over his 
role in weapon manufacturing.”,
'Stark dedicates himself to dismantling the weapons his company has sold 
illegally, using technology and intelligence.',
'Using the drone technology, Stark develops non-lethal weapons and 
countermeasures against illicit arms dealing.',
'Stark confronts Stane, who reveals his grander ambitions to monopolize global 
armaments.',
'Stane, having developed his own advanced weaponry, threatens Stark and his 
ideals.',
'Stark employs his drone army and ingenuity in a non-lethal confrontation 
against Stane.',
“With Potts' help, they expose Stane's dealings, leveraging Stark Industries'  
resources to do so.”,
'Stark transforms Stark Industries into a force for global peacekeeping, 
maintaining a private life while secretly advising on threats.'
]
\end{verbatim}
\end{tcolorbox}
\caption{The generated alternate timeline for Iron Man. See Section \ref{sec:branching} for how the new timeline is generated.}
\label{tab:new_story}  
    \end{tabular}     
\end{table*}


\begin{table*}
\begin{tabular}{p{6.25in}}
\begin{tcolorbox}
\begin{verbatim}
You've decided to go to Afghanistan for the demonstration of the Jericho missile, 
your company's latest and most advanced weapon system. The demonstration goes 
off without a hitch, showcasing the devastating power of what you've created. 
However, the trip takes a dark turn when your convoy is ambushed. The attackers  
are relentless, and amidst the chaos, you're critically wounded, a piece of  
shrapnel lodging near your heart.

In the aftermath, you find yourself captured and imprisoned by the terrorist  
group known as the Ten Rings. They're well aware of who you are and the deadly   
prowess of your weapons. They have a simple demand: build them a Jericho missile.  
But the irony of your situation doesn't escape you – the very weapons you've 
profited from are now a direct threat to your life, and potentially countless
others.

Now, imprisoned by the Ten Rings in a cave in Afghanistan, you're faced with the 
gravest challenge yet. Your captors want a weapon, but you see an opportunity
- not just for escape, but to make a statement against those who'd use your  
creations for terror. You must decide how to use your genius and resources at  
hand not just to survive, but to fight back.

>> Build a suit of armor

>> Sabotage the missile
\end{verbatim}
\end{tcolorbox}
\caption{The new narration for the first alternate decision made for Iron Man. See Section \ref{sec:narrating} for how the new narration is generated.}
\label{tab:narration} 
    \end{tabular} 
    
\end{table*}

\newpage

\begin{table*}
\begin{minipage}{\textwidth}
\section{Prompt Examples}
\label{app:prompt_example}
\subsection{Vanilla Prompt}
\textbf{Extract Events from Plot:}
\begin{tcolorbox}
\begin{verbatim}
(Plot) Summarize the plot above into a list of n events
\end{verbatim}
\end{tcolorbox}

\textbf{Get decisions from each event:}
\begin{tcolorbox}
\begin{verbatim}
(Events) For each event in the event list above, what are the original decision 
and the alternate decision that would lead the main character to a different 
storyline?
\end{verbatim}
\end{tcolorbox}

\textbf{Write new storylines:}
\begin{tcolorbox}
\begin{verbatim}
(Events) Given the event list above, imagine a new storyline where at event x, the 
main character makes the decision x instead of y
\end{verbatim}
\end{tcolorbox}
\end{minipage}
\end{table*}

\begin{table*}
\begin{minipage}{\textwidth}
\subsection{\whatif~Prompts}
The following are all of the prompts we use for the system, in order but without the JSON schema we ask it to use. The JSON schemas for each prompt can be found in Appendix \ref{app:json} and would be filled in via the \Verb|JSON_SCHEMA| variable.
These are the prompts for generating the tree (plot-to-tree), extracting key events, generating meta-prompts, writing a new storyline, and generating the final game narration, respectively.

\begin{tabular}{p{6.25in}}
\begin{tcolorbox}
\begin{verbatim}

{  
   “role”: “system”, 
   “content”: “# You are a helpful fiction writer assistant.”  
},
{
  “role”: “user”, 
  “content”: f“{plot}\n
  Summarize the plot above into a plot tree of 
  {'at most 6' if num_nodes == '' else num_nodes}
   nodes with each node containing the state and goal of {char_name}, and the key 
  decision that propels the story forward. Each edge should contain a list of 
  events that lead {char_name} to the state of next node. Also, Given the same 
   state and goal of {char_name}, imagine an alternate decision that would have led 
  {char_name} to a different storyline. Output in JSON format with schema: 
  {JSON_SCHEMA}. Make sure that all important plot points are included in 
  'edgeEvents' but not in 'state'”
}
\end{verbatim}
\end{tcolorbox}
\caption{Prompt for generating a tree from the plot (plot-to-tree).}
    \end{tabular}
    
\end{minipage}
\end{table*}

\begin{table*}

    \begin{tabular}{p{6.25in}}
    \begin{tcolorbox}
         \begin{verbatim}
{
  “role”: “system”, 
  “content”: f“Here are some definitions in the context of three-act story
  structure: The inciting incident is an event that pulls the protagonist
  out of their normal world and into the main action of the story. It is
  the turning point between Act One and Act Two. The crisis is the moment  
  when the protagonist faces their greatest challenge or obstacle, leading  
  directly to the climax of the story. It is the turning point between Act Two 
  and Act Three. The climax is the climactic confrontation in which the hero 
  faces a point of no return: they must either prevail or perish. It occurs in   
  Act Three and should have the peak tension of the story. You will be given a
  list of events from a movie plot. Your task is to identify the inciting  
  incident, crisis, and climax. Output in JSON format with schema:
  {JSON_SCHEMA}.”
},
{
  “role”: “user”, “content”: f“{events}”  
},
\end{verbatim}
\end{tcolorbox}
 \caption{Prompt for extracting key events.}
 \label{tab:three-act}
    \end{tabular}
    
\end{table*}

\begin{table*}

    \begin{tabular}{p{6.25in}}
    \begin{tcolorbox}
         \begin{verbatim}
{
  “role”: “system”, 
  “content”: “# You are an expert in prompting ChatGPT.”
},
{
 “role”: “user”, 
 “content”: f“Original storyline:{all_events}\n
 Write a prompt for ChatGPT with following requirements:
 1. Ask to use the original storyline as a reference to write an alternate  
    storyline that branches out at event {branching_event} if {char_name}
    {storyline[f'node_{branching_node}']['alternate_decision']} instead of 
    {storyline[f'node_{branching_node}']['decision']}.
 2. Provide 5 thought-provoking concrete guiding questions as potential 
    directions to explore that expand the following:\n
    a. How would alternate decision change or replace {mpp}?
    b. How would {char_name} make key decisions that overcome new challenges
       and propel the story forward?
 3. Describe what an ideal alternate storyline should look like.
 4. Ask to output the alternate storyline as a list of 
    {(len(storyline) - branching_node + 1) * 3} events that has 
    {storyline[f'node_{branching_node}']['alternate_decision']} as the first
    event.\n
 Output the prompt with the following JSON schema: {JSON_SCHEMA}”
}
  \end{verbatim}
  \end{tcolorbox}
  \caption{The prompt for generating meta-prompts.}
    \end{tabular}
    
\end{table*}

\begin{table*}

    \begin{tabular}{p{6.25in}}
    \begin{tcolorbox}
         \begin{verbatim}
{
  “role”: “system”,
  “content”: “# You are a helpful fiction writer assistant.”
},
{
    “role”: “user”, 
    “content”: f“Original storyline:\n
    {all_events}\n\n
    {prompt}\n
    Output in JSON format with schema: {JSON_SCHEMA}.”
}
             \end{verbatim}
             \end{tcolorbox}
             \caption{The prompt for writing the new storyline}
    \end{tabular}
    
\end{table*}

\begin{table*}

    \begin{tabular}{p{6.25in}}
    \begin{tcolorbox}
         \begin{verbatim}
{
   “role”: “system”, “content”: f“# You are  writing a Choose Your Own Adventure  
   style interactive fiction game in which the player is {char_name}.
   You will be given a list of events, the resulting state and goal of the  
   character, and two decisions.
   Do the following:
        1. Narrate each event in a paragraph. You should never mention 
           {char_name} but always use the second-person perspective.
        2. Seemlessly transit to the state and goal of the player.
        3. Provide two button-text reflects the two decisions.
   Output in JSON with schema: {JSON_SCHEMA}”
},
{
  “role”: “user”, 
  “content”: f“{node}”
}
\end{verbatim}
\end{tcolorbox}
\caption{The prompt for generating the final game narration.}
    \end{tabular}
    
\end{table*}

\begin{table*}
\begin{minipage}{\textwidth}
\section{JSON Schema}
\label{app:json}
The following are all of the JSON that are used in the prompts above.
They are for generating the tree (plot-to-tree), extracting key events, generating meta-prompts, writing a new storyline, generating the final game narration.

    \begin{tabular}{p{6.25in}}
    \begin{tcolorbox}
    \begin{verbatim}
{{
“node_1”: {{
          “state”: “<initial state of {char_name}>”,    /* The initial state of the 
              main character. This should NOT contain any important plot point. */
          “goal”: “<goal of {char_name} given the current state>”,    /* The goal
              the main character is attempting to reach in the current state. This  
           should starts with 'To ...' */
          “decision”: “<key decision taken by {char_name} that propels the story  
               forward>”,   /* The key decision taken by {char_name} given the state  
           and goal, starting with '{char_name} decides to ...'. */
         “edgeEvents”: [     /* List of specific events resulting from the key
           decision and leading up to the state of next node. Each event 
           should be a complete sentence with all involved characters */
           “<repeat key decision taken by {char_name} that propels the story  
           forward, starting with '{char_name} decides to ...'>”,
             “<event resulting from the key decision and leading to next state>”,
           “<next state of {char_name} resulting from the previous events>”
           ],
         “alternate_decision”: “<an alternate decision {char_name} could have  
           made given the same state and goal that would have led to a   
           different storyline, starting with '{char_name} decides to ...'>”
        }},
    
“node_2”: {{
       “state”: “<state of the character resulting from the previous node's 
           edgeEvents>”,   /* The current state of the main character, resulted  
          from the previous node's edgeEvents. This should NOT contain any  
          important plot point. */
          “goal”: “<goal of the character given the current state>”,    /* The goal
          the main character is attempting to reach in the current state.
          This should starts with 'To ...' */
          “decision”: “<key decision taken by {char_name} that propels the story 
               forward>”,    /* The key decision taken by {char_name} given the state  
           and goal, starting with '{char_name} decides to ...' */
        “edgeEvents”: [   /* List of specific events resulting from the key
           decision and leading up to the state of next node. Each event  
           should be a complete sentence with all involved characters */
           “<repeat key decision taken by {char_name} that propels the story 
           forward, starting with '{char_name} decides to ...'>”,
            “<event resulting from the key decision and leading to next state>”,
           “<next state of {char_name} resulting from the previous events>”
          ],
           “alternate_decision”: “<an alternate decision {char_name} could have made 
           given the same state and goal that would have led to a different   
           storyline, starting with '{char_name} decides to ...'>”
        }},
\end{verbatim}
\end{tcolorbox}
    \end{tabular}
    \end{minipage}
\end{table*}

\begin{table*}

    \begin{tabular}{p{6.25in}}
    \begin{tcolorbox}
\begin{verbatim}
/* ...continue for all {num_nodes} nodes... */

“node_n”: {{     /* n is the total number of nodes */
        “state”: “<state of the character resulting from previous node's
               edgeEvents>”,    /* The current state of the main character, resulted 
           from the previous node's edgeEvents. This should NOT contain any 
           important plot point. */
            “goal”: “<final character goal given the current state>”,    /* The goal 
              the main character is attempting to reach in the final state. should  
           This starts with 'To ...' */
          “decision”: “<key decision taken by {char_name} that propels the story 
               forward>”,    /* The key decision taken by {char_name} given the state  
           and goal, starting with '{char_name} decides to ...' */
             “edgeEvents”: [     /* List of final events resulting from the key decision  
              and leading to the end of the story. Each event should be a complete  
           sentence with all involved characters */
            “<repeat key decision taken by {char_name} leading to end of story,  
           starting with '{char_name} decides to ...'>”,
                “<event resulting from the key decision and leading to end of story>”,
           “<final state of {char_name} resulting from the previous events>”
           ],
         “alternate_decision”: “<an alternate decision {char_name} could have  
               made given the same state and goal that would have led to a different 
           storyline, starting with '{char_name} decides to ...'>”
         }}
}}
\end{verbatim}
\end{tcolorbox}
\caption{Plot to Tree JSON}
    \end{tabular}
    
\end{table*}

\textbf{}
\begin{table*}

    \begin{tabular}{p{6.25in}}
    \begin{tcolorbox}
         \begin{verbatim}
{
“inciting_incident”: {
    “eventId”: “the event number”,
    “event”: “the event corresponding to the inciting incident”
    },
“crisis”: {
    “eventId”: “the event number”,
    “event”: “the event corresponding to the crisis”
    },
“climax”: {
    “eventId”: “the event number”,
    “event”: “the event corresponding to the climax”
    }
}
\end{verbatim}
\end{tcolorbox}
\caption{Extract Key Events JSON}
    \end{tabular}
    
\end{table*}

\begin{table*}

    \begin{tabular}{p{6.25in}}
    \begin{tcolorbox}
         \begin{verbatim}
{{
“branching_event_number”: branching_event,
“original_decision” : storyline[f“node_{branching_node}”]['decision'],
“alternate_decision” : 
     storyline[f“node_{branching_node}”]['alternate_decision'],
“new_story_length”: (len(storyline) - branching_node) * 3,
“major_plot_points” : mpp,
“prompt”: f“TODO: <a prompt for ChatGPT for every branching point above with 
  following requirements:
       1. Ask to use the original storyline as a reference to write an alternate 
         storyline that branches out at event {branching_event} if {char_name} 
         {storyline[f'node_{branching_node}']['alternate_decision']} instead
         of {storyline[f'node_{branching_node}']['decision']}.
      2. Provide 5 thought-provoking concrete guiding questions as potential 
         directions to explore that expand the following:\n
            a. How would alternate decision change or replace {mpp}?
            b. How would {char_name} make key decisions that overcome new  
               challenges and propel the story forward?
      3. Describe what an ideal alternate storyline should look like.
      4. Ask to output the alternate storyline as a list of 
         {(len(storyline) - branching_node + 1) * 3} events that has 
         {storyline[f'node_{branching_node}']['alternate_decision']} as the 
         first event>”
}}
\end{verbatim}
\end{tcolorbox}
\caption{Generate Prompt JSON}
    \end{tabular}
    
\end{table*}

\begin{table*}

    \begin{tabular}{p{6.25in}}
    \begin{tcolorbox}
         \begin{verbatim}
{
“events”: {
           “event number”: “an event in the new storyline”
          }
}
\end{verbatim}
\end{tcolorbox}
\caption{Write New Storyline JSON}
    \end{tabular}
    
\end{table*}

\begin{table*}

    \begin{tabular}{p{6.25in}}
    \begin{tcolorbox}
         \begin{verbatim}

{
“paragraphs”: “<Narrate the three events as three short paragraphs using 
   second-person perspective, then transit to the state and goal of the player. 
   Use newline characters between paragraphs>”,
“button_text_1”: “<short button text for original decision>”,
“button_text_2”: “<short button text for alternate decision>”
}
\end{verbatim}
\end{tcolorbox}
\caption{Narration JSON}
    \end{tabular}
    
\end{table*}
\end{document}